# Instance-sensitive Fully Convolutional Networks

Jifeng Dai[1], Kaiming He[1], Yi Li[2*], Shaoqing Ren[3*], Jian Sun[1]

[1]Microsoft Research, [2]Tsinghua University,
[3]University of Science and Technology of China

**Abstract** Fully convolutional networks (FCNs) have been proven very successful for semantic segmentation, but the FCN outputs are unaware of object instances. In this paper, we develop FCNs that are capable of proposing instance-level segment candidates. In contrast to the previous FCN that generates one score map, our FCN is designed to compute a small set of *instance-sensitive* score maps, each of which is the outcome of a pixel-wise classifier of a relative position to instances. On top of these instance-sensitive score maps, a simple assembling module is able to output instance candidate at each position. In contrast to the recent DeepMask method for segmenting instances, our method does not have any high-dimensional layer related to the mask resolution, but instead exploits image local coherence for estimating instances. We present competitive results of instance segment proposal on both PASCAL VOC and MS COCO.

## 1 Introduction

Fully convolutional networks (FCN) [1] have been proven an effective end-to-end solution to semantic image segmentation. An FCN produces a score map of a size proportional to the input image, where every pixel represents a classifier of objects. Despite good accuracy and ease of usage, FCNs are not directly applicable for producing instance segments (Fig. 1 (top)). Previous instance semantic segmentation methods (*e.g.*, [2,3,4,5]) in general resorted to off-the-shelf segment proposal methods (*e.g.*, [6,7]).

In this paper, we develop an end-to-end fully convolutional network that is capable of segmenting candidate instances. Like the FCN in [1], in our method *every pixel still represents a classifier*; but unlike an FCN that generates one score map (for one object category), our method computes a set of *instance-sensitive* score maps, where each pixel is a classifier of *relative positions* to an object instance (Fig. 1 (bottom)). For example, with a 3×3 regular grid depicting relative positions, we produce a set of 9 score maps in which, *e.g.*, the map #6 in Fig. 1 has high scores on the "right side" of object instances. With this set of score maps, we are able to generate an object instance segment in each sliding window by *assembling* the output from the score maps. This procedure enables a fully convolutional way of producing segment instances.

---

* This work was done when Yi Li and Shaoqing Ren were interns at Microsoft Research.

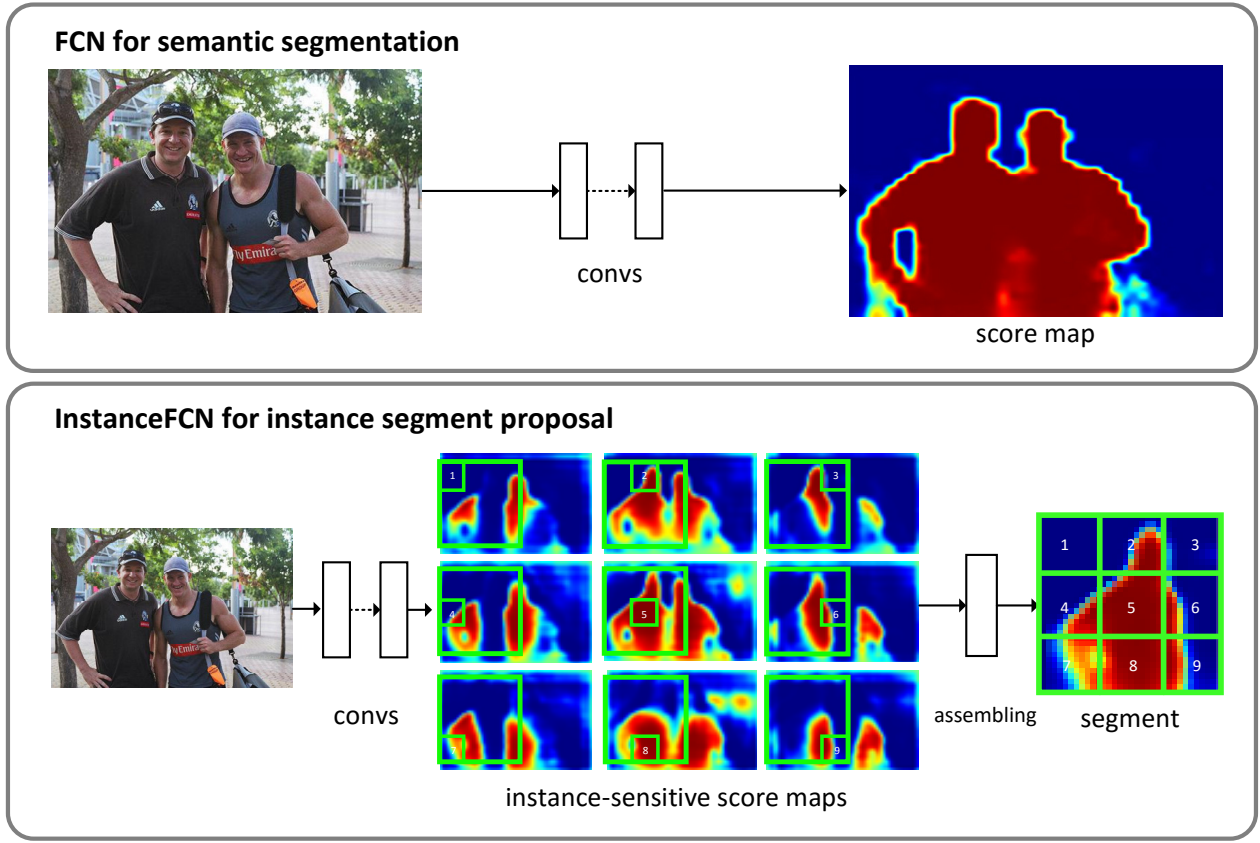

**Figure 1.** Methodological comparisons between: (**top**) FCN [1] for semantic segmentation; (**bottom**) our InstanceFCN for instance segment proposal.

Most related to our method, *DeepMask* [8] is an instance segment proposal method driven by convolutional networks. DeepMask learns a function that maps an image sliding window to an $m^2$-d vector representing an $m\times m$-resolution mask (*e.g.*, $m = 56$). This is computed by an $m^2$-d *fully-connected* (*fc*) layer. See Fig. 2. Even though DeepMask can be implemented in a fully convolutional way (as at inference time in [8]) by recasting this *fc* layer into a convolutional layer with $m^2$-d outputs, it fundamentally differs from the FCNs in [1] where each output pixel is a *low-dimensional* classifier. Unlike DeepMask, our method has no layer whose size is related to the mask size $m^2$, and each pixel in our method is a low-dimensional classifier. This is made possible by exploiting *local coherence* [9] of natural images for generating per-window pixel-wise predictions. We will discuss and compare with DeepMask in depth.

On the PASCAL VOC [10] and MS COCO [11] benchmarks, our method yields compelling instance segment proposal results, comparing favorably with a series of proposal methods [6,12,8]. Thanks to the small size of the layer for predicting masks, our model trained on the small PASCAL VOC dataset exhibits good accuracy with less risk of overfitting. In addition, our system also shows competitive results for instance semantic segmentation when used with downstream classifiers. Our method, dubbed *InstanceFCN*, shows that segmenting instances can still be addressed by the FCN fashion in [1], filling a missing piece among the broad applications of FCNs.

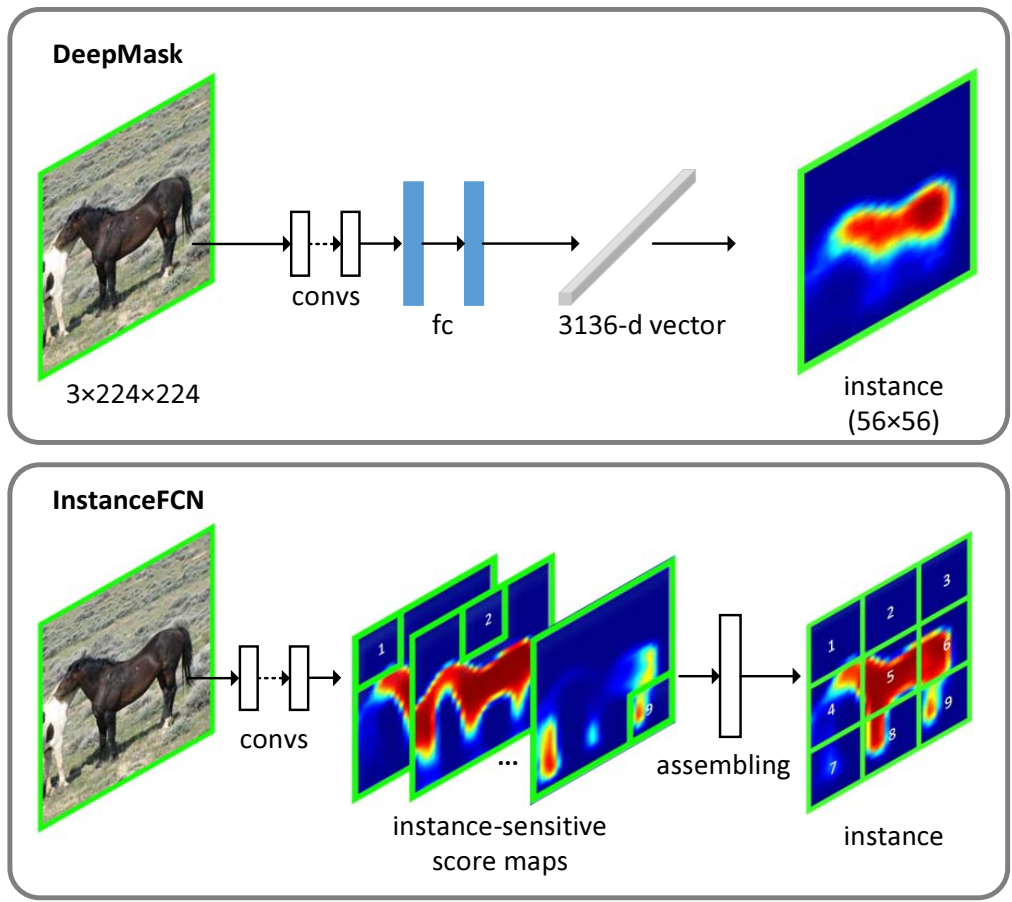

**Figure 2.** Methodological comparisons between DeepMask [8] and InstanceFCN for instance segment proposal. DeepMask uses a high-dimensional $m^2$-d *fc* layer to generate an instance, *e.g.*, $m = 56$ and $m^2 = 3136$. Our network has no any $m^2$-d layer.

## 2 Related Work

The general concept of fully convolutional models dates back to at least two decades ago [13]. For convolutional neural networks (CNNs) [14,15], a sliding window (or referred to as a patch or crop) is not necessarily run on the image domain but instead is run on a feature map, which can be recast into convolutional filters on that feature map. These fully convolutional models are naturally applicable for image restoration problems, such as denoising [16], super-resolution [17], and others [18], where each output pixel is a real-number regressor of intensity values.

Recently FCNs [1] have shown compelling quality and efficiency for semantic segmentation. In [1], each output pixel is a classifier corresponding to the receptive field of the network. The networks can thus be trained end-to-end, pixel-to-pixel, given the category-wise semantic segmentation annotation. But this method can not distinguish object instances (Fig. 1).

Operated fully convolutionally, the Region Proposal Network (RPN) in Faster R-CNN [19] is developed for proposing box-level instances. In an RPN, each pixel of the output map represents a bounding box regressor and an objectness classifier. The RPN does not generate mask-level proposals. In [20], the RPN boxes are used for regressing segmentation masks, conducted by an *fc* layer on Region-of-Interest (RoI) pooling features [21].

## 3 Instance-sensitive FCNs for Segment Proposal

### 3.1 From FCN to InstanceFCN

Although the original FCN [1] for semantic segmentation produces no explicit instance, we can still think of some special cases in which such an FCN can do *a good job* generating an instance. Let's consider an image that contains only one object instance. In this case, the original FCN can produce a good mask about this object category, and because there is only one instance, this is also a good mask about this object instance. In this procedure, the FCN does not have any pre-define filters that are dependent on the mask resolution/size (say, $m \times m$).

Next let's consider an image that contains two object instances that are close to each other (Fig. 1(top)). Although now the FCN output (Fig. 1(top)) does not distinguish the two instances, we notice that the output is indeed *reusable for most pixels*, except for those where one object is conjunct the other — *e.g.*, when the "right side" of the left instance is conjunct the "left side" of the right instance (Fig. 1). If we can discriminate "right side" from "left side", we can still rely on FCN-like score maps to generate instances.

**Instance-sensitive score maps**

The above analysis motivates us to introduce the concept of *relative positions* into FCNs. Ideally, relative positions are with respect to object instances, such as the "right side" of an object or the "left side" of an object. In contrast to the original FCN [1] where each output pixel is a classifier of an object category, we propose an FCN where each output pixel is *a classifier of relative positions of instances*. For example, for the #4 score map in Fig. 1 (bottom), each pixel is a classifier of being or not being "left side" of an instance.

In our practice, we define the relative positions using a $k \times k$ (*e.g.*, $k = 3$) regular grid on a square sliding window (Fig. 1 (bottom)). This leads to a set of $k^2$ (*e.g.*, 9) score maps which are our FCN outputs. We call them *instance-sensitive score maps*. The network architecture for producing these score maps can be trained end-to-end, with the help of the following module.

**Instance assembling module**

The instance-sensitive score maps have not yet produced object instances. But we can simply assemble instances from these maps. We slide a window of resolution $m \times m$ on the set of instance-sensitive score maps (Fig. 1 (bottom)). In this sliding window, each $\frac{m}{k} \times \frac{m}{k}$ sub-window directly copies values from the same sub-window in the corresponding score map. The $k^2$ sub-windows are then put together (according to their relative positions) to assemble a new window of resolution $m \times m$. This is the instance assembled from this sliding window.

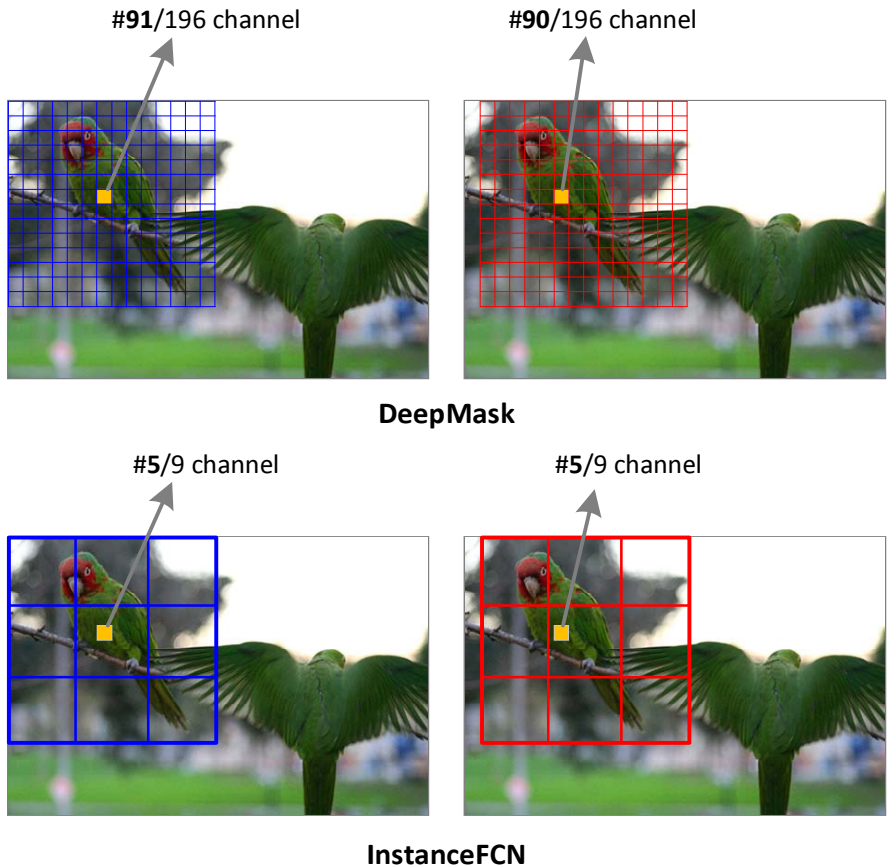

**Figure 3.** Our method can exploit image *local coherence*. For a window shifted by one small step (from blue to red), our method can reuse the same prediction from the same score map at that pixel. This is not the case if the masks are produced by a sliding $m^2$-dimensional *fc* layer (for illustration $m = 14$ in this figure).

This instance assembling module is adopted for both training and inference. During training, this model generates instances from sparsely sampled sliding windows, which are compared to the ground truth. During inference, we densely slide a window on the feature maps to predict an instance segment at each position. More details are in the algorithm section.

We remark that the assembling module is *the only component* in our architecture that involves the mask resolution $m \times m$. Nevertheless, the assembling module has no network parameter to be learned. It is inexpensive because it only has copy-and-paste operations. This module impacts training as it is used for computing the loss function.

### 3.2   Local Coherence

Next we analyze our method from the perspective of *local coherence* [9]. By local coherence we mean that for a pixel in a natural image, its prediction is most likely the same when evaluated in two neighboring windows. One does not need to completely re-compute the predictions when a window is shifted by a small step.

The local coherence property has been exploited by our method. For a window that slides by one stride (Fig. 3 (bottom)), the same pixel in the image coordinate system will have the same prediction because it is copied from the same score map (except for a few pixels near the partitioning of relative positions). This allows us to conserve a large number of parameters when the mask resolution $m^2$ is high.

This is in contrast to DeepMask's [8] mechanism which is based on a "sliding *fc* layer" (Fig. 3 (top)). In DeepMask, when the window is shifted by one stride, the same pixel in the image coordinate system is predicted by two different channels of the *fc* layer, as shown in Fig. 3 (top). So the prediction of this pixel is in general not the same when evaluated in two neighboring windows.

By exploiting local coherence, our network layers' sizes and dimensions are all independent of the mask resolution $m \times m$, in contrast to DeepMask. This not only reduces the computational cost of the mask prediction layers, but more importantly, reduces the number of parameters required for mask regression, leading to less risk of overfitting on small datasets such as PASCAL VOC. In the experiment section we show that our mask prediction layer can have hundreds times fewer parameters than DeepMask.

### 3.3 Algorithm and Implementation

Next we describe the network architecture, training algorithm, and inference algorithm of our method.

**Network architecture.** As common practice, we use the VGG-16 network [22] pre-trained on ImageNet [23] as the feature extractor. The 13 convolutional layers in VGG-16 are applied fully convolutionally on an input image of arbitrary size. We follow the practice in [24] to reduce the network stride and increase feature map resolution: the max pooling layer $pool_4$ (between $conv_{4\_3}$ and $conv_{5\_1}$) is modified to have a stride of 1 instead of 2, and accordingly the filters in $conv_{5\_1}$ to $conv_{5\_3}$ are adjusted by the "hole algorithm" [24]. Using this modified VGG network, the effective stride of the $conv_{5\_3}$ feature map is $s = 8$ pixels *w.r.t.* the input image. We note that this reduced stride directly determines the resolutions of our score maps from which our masks are copied and assembled.

On top of the feature map, there are two fully convolutional branches, one for estimating segment instances and the other for scoring the instances. For the first branch, we adopt a 1×1 512-d convolutional layer (with ReLU [25]) to transform the features, and then use a 3×3 convolutional layer to generate a set of instance-sensitive score maps. With a $k \times k$ regular grid for describing relative positions, this last convolutional layer has $k^2$ output channels corresponding to the set of $k^2$ instance-sensitive score maps. See the top branch in Fig. 4. On top of these score maps, an assembling module is used to generate object instances in a sliding window of a resolution $m \times m$. We use $m = 21$ pixels (on the feature map with a stride of 8).

For the second branch of scoring instances (bottom in Fig. 4), we use a 3×3 512-d convolutional layer (with ReLU) followed by a 1×1 convolutional layer. This 1×1 layer is a per-pixel logistic regression for classifying instance/not-instance of the sliding window centered at this pixel. The output of this branch is thus an objectness score map (Fig. 4 (bottom)), in which one score corresponds to one sliding window that generates one instance.

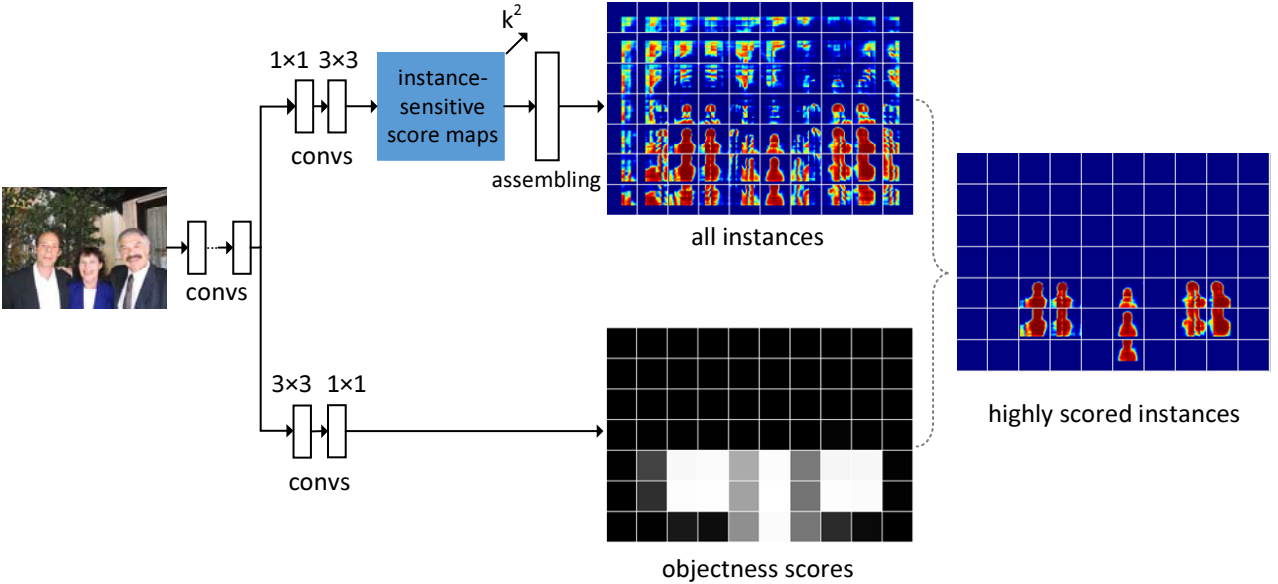

**Figure 4.** Details of the InstanceFCN architecture. On the top is a fully convolutional branch for generating $k^2$ instance-sensitive score maps, followed by the assembling module that outputs instances. On the bottom is a fully convolutional branch for predicting the objectness score of each window. The highly scored output instances are on the right. In this figure, the objectness map and the "all instances" map have been sub-sampled for the purpose of illustration.

**Training.** Our network is trained end-to-end. We adopt the *image-centric* strategy in [21,19]. The forward pass computes the set of instance-sensitive score maps and the objectness score map. After that, a set of 256 sliding windows are randomly sampled [21,19], and the instances are only assembled from these 256 windows for computing the loss function. The loss function is defined as:

$$\sum_i (\mathcal{L}(p_i, p_i^*) + \sum_j \mathcal{L}(S_{i,j}, S_{i,j}^*)). \qquad (1)$$

Here $i$ is the index of a sampled window, $p_i$ is the predicted objectness score of the instance in this window, and $p_i^*$ is 1 if this window is a positive sample and 0 if a negative sample. $S_i$ is the assembled segment instance in this window, $S_i^*$ is the ground truth segment instance, and $j$ is the pixel index in the window. $\mathcal{L}$ is the logistic regression loss. We use the definition of positive/negative samples in [8], and the 256 sampled windows have a positive/negative sampling ratio of 1:1 [19].

Our model accepts images of arbitrary size as input. We follow the scale jittering in [26] for training: a training image is resized such that its shorter side is randomly sampled from $600 \times 1.5^{\{-4,-3,-2,-1,0,1\}}$ pixels. We use Stochastic Gradient Descent (SGD) as the solver. A total of 40k iterations are performed, with a learning rate of 0.001 for the first 32k and 0.0001 for the last 8k. We perform training with an 8-GPU implementation, where each GPU holds 1 image with 256 sampled windows (so the effective mini-batch size is 8 images). The weight decay is 0.0005 and the momentum is 0.9. The first thirteen convolutional layers are initialized by the ImageNet pre-trained VGG-16 [22], and the extra convolutional layers are randomly initialized from a Gaussian distribution with zero mean and standard derivation of 0.01.

**Table 1.** Ablation experiments on the numbers of instance-sensitive score maps (*i.e.*, # of relative positions, $k^2$), evaluated on the PASCAL VOC 2012 validation set.

| $k^2$ | AR@10 (%) | AR@100 (%) | AR@1000 (%) |
|---|---|---|---|
| $3^2$ | 38.3 | 49.2 | 52.1 |
| $5^2$ | <u>38.9</u> | <u>49.7</u> | 52.6 |
| $7^2$ | 38.8 | <u>49.7</u> | <u>52.7</u> |

**Inference.** A forward pass of the network is run on the input image, generating the instance-sensitive score maps and the objectness score map. The assembling module then applies densely sliding windows on these maps to produce a segment instance at each position. Each instance is associated with a score from the objectness score map. To handle multiple scales, we resize the shorter side of images to $600 \times 1.5^{\{-4,-3,-2,-1,0,1\}}$ pixels, and compute all instances at each scale. It takes totally 1.5 seconds evaluating an images on a K40 GPU.

For each output segment, we truncate the values to form a binary mask. Then we adopt non-maximum suppression (NMS) to generate the final set of segment proposals. The NMS is based on the objectness scores and the box-level IoU given by the tight bounding boxes of the binary masks. We use a threshold of 0.8 for the NMS. After NMS, the top-$N$ ranked segment proposals are used as the output.

## 4 Experiments

### 4.1 Experiments on PASCAL VOC 2012

We first conduct experiments on PASCAL VOC 2012 [10]. Following [3,4], we use the segmentation annotations from [27], and train the models on the training set and evaluate on the validation set. All segment proposal methods are evaluated by the mask-level intersection-over-union (IoU) between the predicted instances and the ground-truth instances. Following [8], we measure the Average Recall (AR) [28] (between IoU thresholds of 0.5 to 1.0) at a fixed number $N$ of proposals, denoted as "AR@$N$". In [28], the AR metrics have been shown to be more correlated to the detection accuracy (when used with downstream classifiers [2,21]) than traditional metrics for evaluating proposals.

**Ablations on the number of relative positions $k^2$**

Table 1 shows our results using different values of $k^2$. Our method is not sensitive to $k^2$, and can perform well even when $k = 3$. Fig. 5 shows some examples of the instance-sensitive maps and assembled instances for $k = 3$.

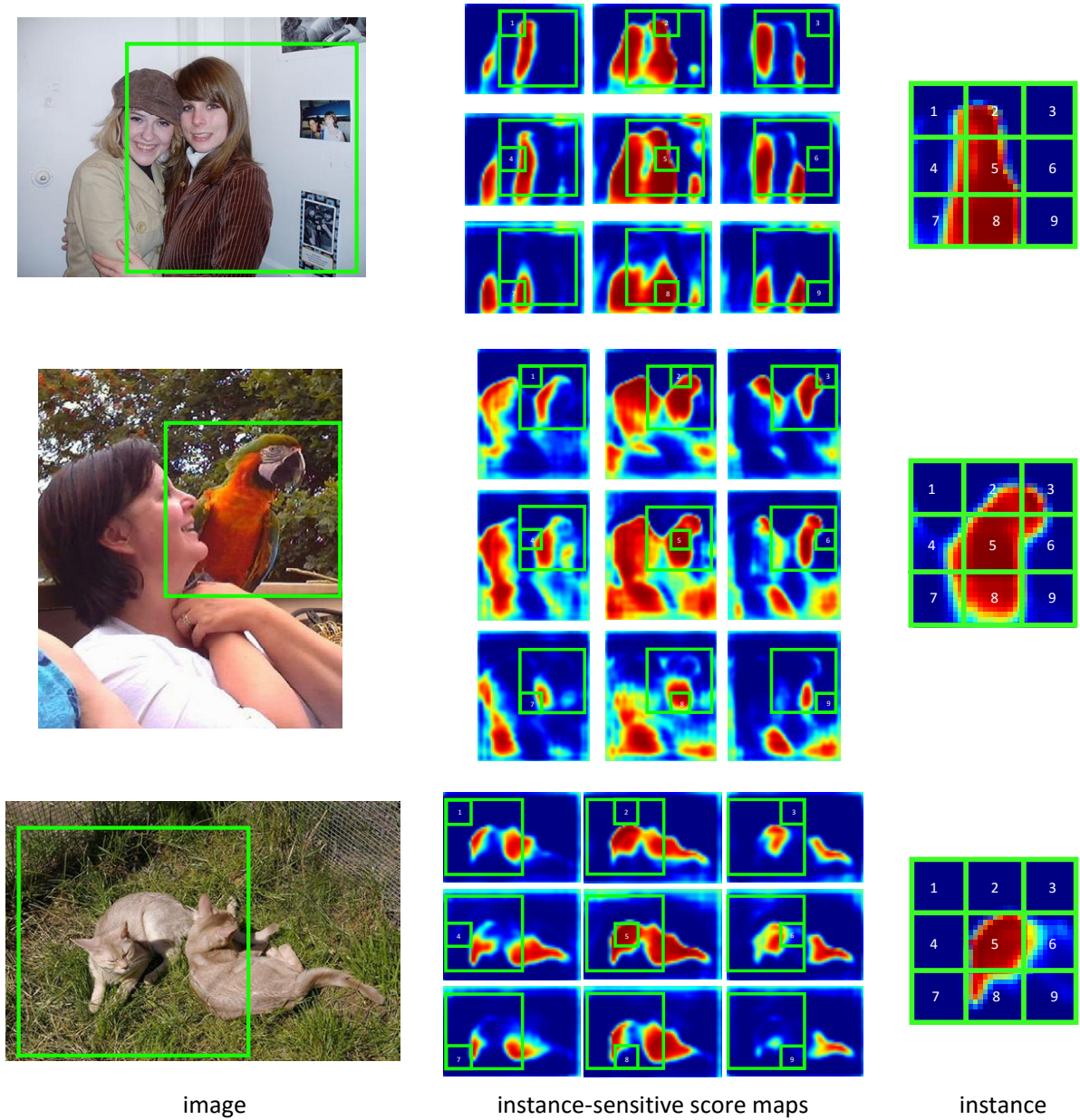

| image | instance-sensitive score maps | instance |

**Figure 5.** Examples of instance-sensitive maps and assembled instances on the PASCAL VOC validation set. For simplicity we only show the cases of $k = 3$ (9 instance-sensitive score maps) in this figure.

Table 1 also shows that our results of $k = 5$ and $k = 7$ are comparable, and are slightly better than the $k = 3$ baseline. Our method enjoys a small gain with a finer division of relative position, but gets saturated around $k = 5$. In the following experiments we use $k = 5$.

**Ablation comparisons with the DeepMask scheme**

For fair comparisons, we implement a DeepMask baseline on PASCAL VOC. Specifically, the network structure is VGG-16 followed by an extra 512-d 1×1 convolutional layer [8], generating a 14×14 feature map as in [8] from a 224×224 image crop. Then a 512-d *fc* layer [8] is applied to this feature map, followed by a $56^2$-d *fc* [8] for generating a 56×56-resolution mask. The two *fc* layers under this setting have 53M parameters[1]. The objectness scoring branch is constructed as in [8]. All other settings are the same as ours for fair comparisons. We refer to this model as **~DeepMask** which means our implementation of DeepMask. This baseline's results are in Table 2.

**Table 2.** Ablation comparisons between ~DeepMask and our method on the PASCAL VOC 2012 validation set. "~DeepMask" is our implementation based on controlled settings (see more descriptions in the main text).

| method | train | test | AR@10 (%) | AR@100 (%) | AR@1000 (%) |
|---|---|---|---|---|---|
| ~DeepMask | crop 224×224 | sliding *fc* | 31.2 | 42.9 | 47.0 |
| ours | crop 224×224 | fully conv. | 37.4 | 48.4 | 51.4 |
| ours | fully conv. | fully conv. | **38.9** | **49.7** | **52.6** |

**Table 3.** Comparisons with state-of-the-art segment proposal methods on the PASCAL VOC 2012 validation set. The results of SS [6] and MCG [12] are from the publicly available code, and the results of MNC [20] is provided by the authors of [20].

| method | AR@10 (%) | AR@100 (%) | AR@1000 (%) |
|---|---|---|---|
| SS [6] | 7.0 | 23.5 | 43.3 |
| MCG [12] | 18.9 | 36.8 | 49.5 |
| ~DeepMask | 31.2 | 42.9 | 47.0 |
| MNC [20] | <u>33.4</u> | <u>48.5</u> | **53.8** |
| ours | **38.9** | **49.7** | <u>52.6</u> |

Table 2 shows the ablation comparisons. As the first variant, we train our model on 224×224 crops as is done in DeepMask. Under this ablative training, our method still outperforms ~DeepMask by healthy margins. When trained on full-size images (Table 2), our result is further improved. The gain from training on full-size images further demonstrates the benefits of our fully convolutional scheme.

It is noteworthy that our method has considerably fewer parameters. Our last $k^2$-d convolutional layer has only 0.1M parameters[2] (all other layers being the same as the DeepMask counterpart). This mask generation layer has only 1/500 of parameters comparing with DeepMask's *fc* layers. Regressing high-dimensional $m \times m$ masks is possible for our method as it exploits local coherence. We also expect fewer parameters to have less risk of overfitting.

**Comparisons with state-of-the-art segment proposal methods**

In Table 3 and Fig. 6 we compare with state-of-the-art segment proposal methods: Selective Search (SS) [6], Multiscale Combinatorial Grouping (MCG) [12], ~DeepMask, and Multi-task Network Cascade (MNC) [20]. MNC is a joint multi-stage cascade method that proposes box-level regions, regresses masks from these regions, and classifies these mask. With a trained MNC, we treat the mask regression outputs as the segment proposals.

---

[1] $512 \times 14 \times 14 \times 512 + 512 \times 56^2 = 53M$
[2] $512 \times 3 \times 3 \times 25 = 0.1M$

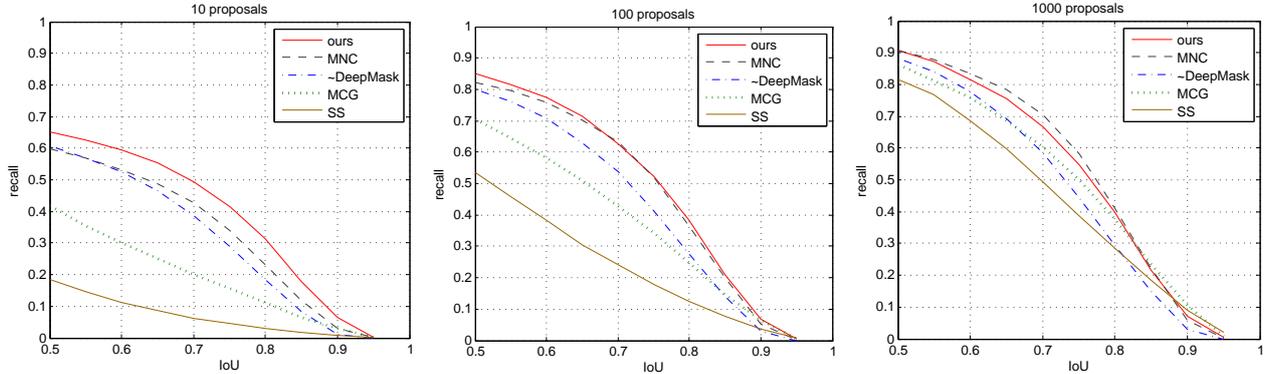

**Figure 6.** Recall *vs.* IoU curves of different segment proposals on the PASCAL VOC 2012 validation set. AR is the area under the curves.

**Table 4.** Semantic instance segmentation on the PASCAL VOC 2012 validation set. All methods are based on VGG-16 except SDS based on AlexNet [15].

| downstream classifier | proposals | mAP@0.5 (%) | mAP@0.7 (%) |
|---|---|---|---|
| SDS [3] | MCG [7] | 49.7 | 25.3 |
| Hypercolumn [4] | MCG [7] | 60.0 | 40.4 |
| CFM [5] | MCG [7] | 60.7 | 39.6 |
| MNC [20] | MNC [20] | **63.5** | <u>41.5</u> |
| MNC [20] | **ours** | <u>61.5</u> | **43.0** |

Table 3 and Fig. 6 show that the CNN-based methods (~DeepMask, MNC, ours) perform better than the bottom-up segmentation methods of SS and MCG. In addition, our method has AR@100 and AR@1000 similar to MNC, but has 5.5% higher AR@10. The mask regression of MNC is done by high-dimensional *fc* layers, in contrast to our fully convolutional fashion.

**Comparisons on Instance Semantic Segmentation**

Next we evaluate the instance semantic segmentation performance when used with downstream category-aware classifiers. Following [3,4], we evaluate mean Average Precision (mAP) using mask-level IoU at threshold of 0.5 and 0.7. In Table 4 we compare with: SDS [3], Hypercolumn [4], CFM [5], and MNC [20]. We use MNC's stage 3 as our classifier structure, which is similar to Fast R-CNN [21] except that its RoI pooling layer is replaced with an RoI masking layer that generates features from the segment proposals. We adopt a two-step training: first train our model for proposing segments and then train the classifier with the given proposals. Our method uses $N = 300$ proposals in this comparison.

Table 4 shows that among all the competitors our method has the highest mAP@0.7 score of 43.0%, which is 1.5% better than the closest competitor. Our method has the second best mAP@0.5, lower than MNC. We note that MNC is a joint training algorithm which simultaneously learns proposals and category classifiers. Our result (61.5%) is based on two-step training, and is better than MNC's step-by-step training counterpart (60.2% [20]).

**Table 5.** Comparisons of instance segment proposals on the first 5k images [8] from the MS COCO validation set. DeepMask's results are from [8].

| segment proposals | AR@10 (%) | AR@100 (%) | AR@1000 (%) |
|---|---|---|---|
| GOP [29] | 2.3 | 12.3 | 25.3 |
| Rigor [30] | - | 9.4 | 25.3 |
| SS [6] | 2.5 | 9.5 | 23.0 |
| MCG [7] | 7.7 | 18.6 | 29.9 |
| DeepMask [8] | 12.6 | 24.5 | 33.1 |
| DeepMaskZoom [8] | 12.7 | 26.1 | 36.6 |
| ours | **16.6** | **31.7** | **39.2** |

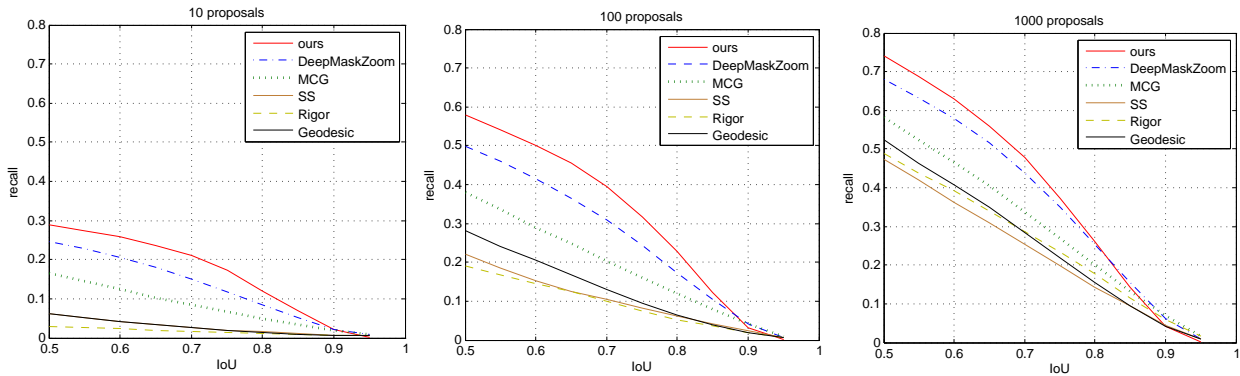

**Figure 7.** Recall *vs.* IoU curves on the first 5k images [8] on the MS COCO validation set. DeepMask's curves are from [8].

### 4.2 Experiments on MS COCO

Finally we evaluate instance segment proposals on the MS COCO benchmark [11]. Following [8], we train our network on the 80k training images and evaluate on the first 5k validation images. The results are in Table 5 (DeepMask's results are reported from [8]). For fair comparisons, we use the same multiple scales used in [8] for training and testing on COCO. Our method has higher AR scores than DeepMask and a DeepMaskZoom variant [8]. Fig. 7 shows the recall *vs.* IoU curves on COCO.

## 5 Conclusion

We have presented InstanceFCN, a fully convolutional scheme for proposing segment instances. It is driven by classifying pixels based on their relative positions, which leads to a set of instance-sensitive score maps. A simple assembling module

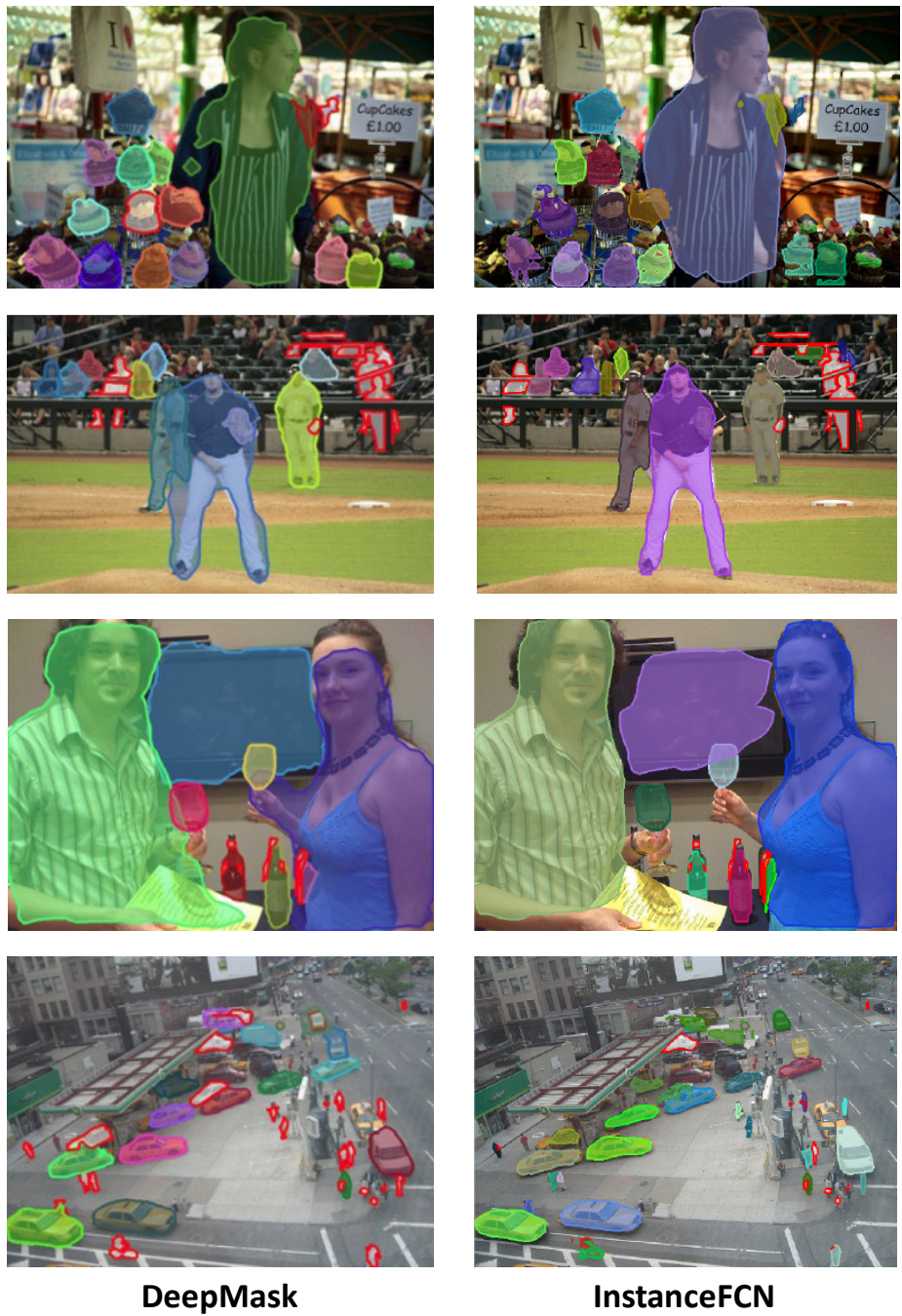

**DeepMask**            **InstanceFCN**

**Figure 8.** Comparisons with DeepMask [8] on the MS COCO validation set. **Left**: DeepMask, taken from the paper of [8]. Proposals with highest IoU to the ground truth are displayed. *The missed ground-truth objects (no proposals with $IoU > 0.5$) are marked by red outlines filled with white.* **Right**: Our results displayed in the same way.

is then able to generate segment instances from these score maps. Our network architecture handles instance segmentation without using any high-dimensional layers that depend on the mask resolution. We expect our novel design of fully convolutional models will further extend the family of FCNs.

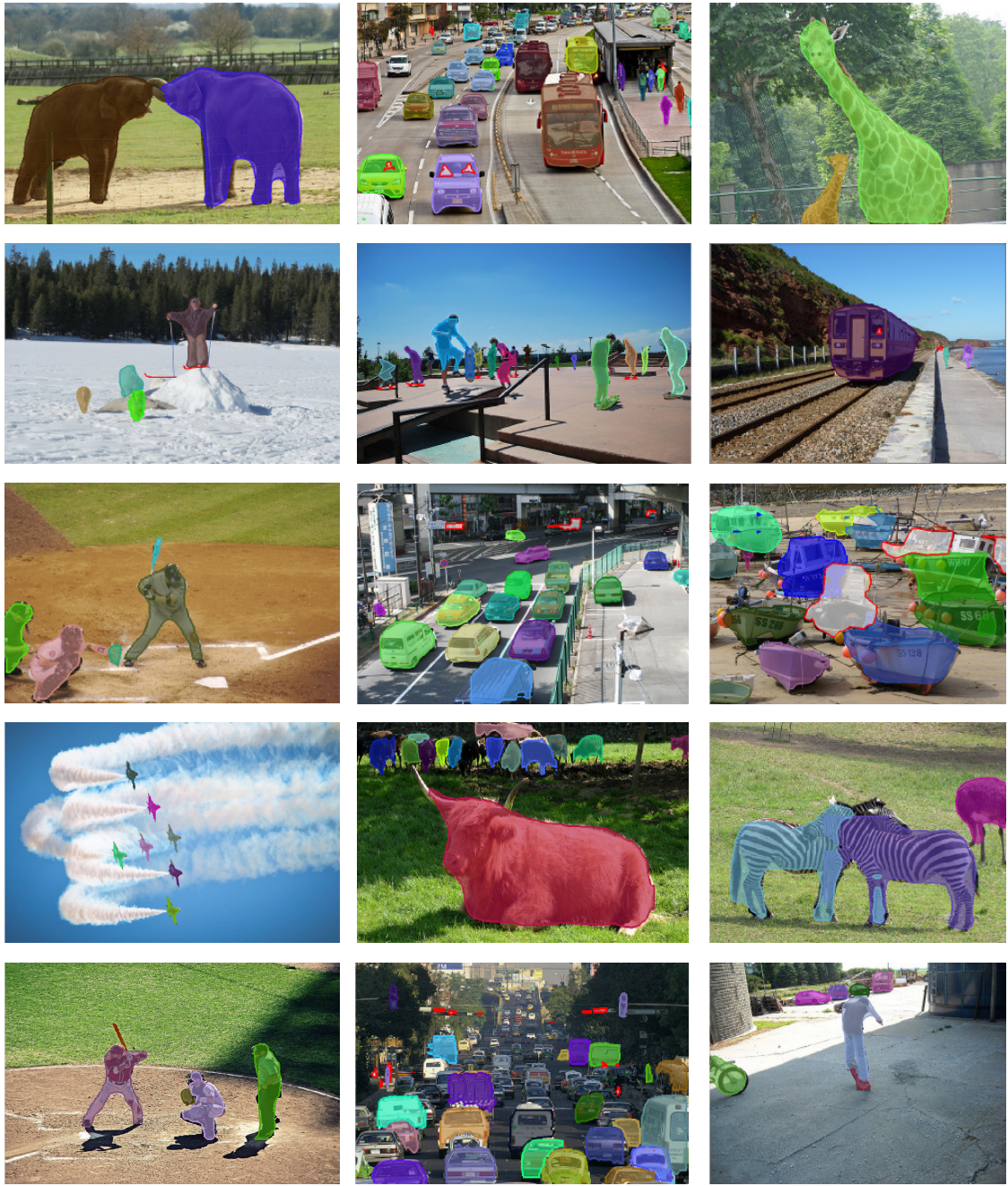

**Figure 9.** More examples of our results on the MS COCO validation set, displayed in the same way of Fig. 8 (*the missed ground-truth objects are marked by red outlines filled with white*).